\documentclass[conference]{IEEEtran}

\usepackage{cite}
\usepackage{amsmath,amssymb,amsfonts}
\usepackage{amsthm}
\usepackage{graphicx}
\usepackage{xcolor}
\usepackage{hyperref}
\usepackage{acronym}
\usepackage{algorithm}
\usepackage{algpseudocode}
\usepackage{enumitem}
\usepackage{needspace}
\acrodef{ai}[AI]{artificial intelligence}
\acrodef{llm}[LLM]{large language model}
\acrodef{gpu}[GPU]{graphics processing unit}
\acrodef{api}[API]{application programming interface}
\acrodef{rest}[REST]{representational state transfer}
\acrodef{react}[ReAct]{reasoning and acting}
\acrodef{cro}[CRO]{Conformal Resource Orchestrator}
\acrodef{e2e}[E2E]{end-to-end}
\acrodef{mle}[MLE]{maximum likelihood estimation}
\acrodef{sla}[SLA]{service level agreement}
\acrodef{dag}[DAG]{directed acyclic graph}
\acrodef{iaas}[IaaS]{infrastructure-as-a-service}
\acrodef{drl}[DRL]{deep reinforcement learning}
\acrodef{moe}[MoE]{mixture-of-experts}
\acrodef{vm}[VM]{virtual machine}
\acrodef{mec}[MEC]{mobile edge computing}
\acrodef{aigc}[AIGC]{AI-generated content}

\newtheorem{theorem}{Theorem}
\newtheorem{remark}{Remark}

\hypersetup{
    colorlinks=true,
    linkcolor=black,
    citecolor=black,
    urlcolor=black
}

\IEEEoverridecommandlockouts

\begin{document}

\title{Uncertainty-Aware Resource Orchestration for Agentic AI Systems}

\author{
    \IEEEauthorblockN{Author Name}
    \IEEEauthorblockA{\textit{Department} \\
        \textit{Organization, Country} \\
        email@example.com}
}

\author{
    \IEEEauthorblockN{Masoud Shokrnezhad}
    \IEEEauthorblockA{\textit{ICTFICIAL Oy, Finland} \\
        masoud.shokrnezhad@ictficial.com}
    \and
    \IEEEauthorblockN{Tarik Taleb}
    \IEEEauthorblockA{\textit{Ruhr University Bochum, Germany} \\
        tarik.taleb@ruhr-uni-bochum.de}
}

\IEEEaftertitletext{\vspace{-1\baselineskip}}

\maketitle

\begin{abstract}
    Agentic artificial intelligence (AI) systems powered by large language models (LLMs) execute as autonomous multi-step pipelines that dynamically invoke tools, integrate observations into memory, and iterate until a task is resolved or a resource budget is exhausted.
    Unlike classical workflows with fixed graphs, the aggregate compute, bandwidth, and end-to-end (E2E) latency demands of such pipelines are unknown at dispatch time: they emerge at runtime from intermediate observations and stochastic token sampling, yielding heavy-tailed, multi-modal random variables.
    This structural unpredictability forces infrastructure orchestrators into a fundamental dilemma: under-provisioning risks mid-execution throttling and deadline violations, while over-provisioning wastes scarce edge or cloud capacity.
    Existing schedulers require a fully known pipeline graph, while edge resource-management methods rely on parametric or stationary demand models, and neither fits the structurally unknown, non-stationary profiles of agentic pipelines.
    We formalize this setting as the uncertainty-aware agentic orchestration problem, where the goal is to reserve, at dispatch time, compute and bandwidth allocations that cover actual pipeline demands, and to verify that the pipeline can meet its deadline, each with a prescribed probability.
    We then propose Conformal Reservation for Orchestration (CRO), an online algorithm that translates a sliding window of past execution traces into dispatch-time resource allocations with finite-sample, distribution-free coverage guarantees via conformal prediction.
    Simulations over heterogeneous task mixes with multi-modal and heavy-tailed profiles show that CRO is the only evaluated method that simultaneously satisfies target probabilistic coverage across all risk levels while, among coverage-feasible methods, attaining the lowest over-reservation, roughly halving the resource waste of the Chebyshev-bound baseline.
\end{abstract}

\begin{IEEEkeywords}
    agentic AI, conformal prediction, resource provisioning, online resource allocation, large language models
\end{IEEEkeywords}

\section{Introduction}\label{sec:introduction}

Agentic \ac{ai} systems powered by \acp{llm} have rapidly matured into autonomous pipelines~\cite{shokrnezhad2026kpi2kvi} that, following a \ac{react}-style loop~\cite{yao2023react}, iteratively select and invoke tools (web searches, code interpreters, database queries, and sub-agent calls), incorporate each observation into memory, and repeat until the task is resolved or a budget is exhausted~\cite{plaat2025agentic, yu2025workflow}.
Critically, the full execution trace, including which tools are invoked, how many steps the pipeline runs, the per-step compute and bandwidth demands, and the \ac{e2e} latency, is not known at task-dispatch time: it emerges at runtime from intermediate observations and stochastic token sampling, rendering the aggregate resource consumption a heavy-tailed, multi-modal\footnote{Throughout this paper, \emph{multi-modal} denotes a probability distribution with multiple modes (peaks), not multimodal foundation models that jointly process text, vision, and other data modalities.}, and task-dependent random variable.
For infrastructure orchestrators, this creates a fundamental challenge: resources must be reserved \emph{before} execution begins, yet under-provisioning risks mid-execution throttling and deadline violations, while over-provisioning wastes scarce edge or cloud capacity.
The central problem we address is therefore how an orchestrator can reserve compute and bandwidth at dispatch time with provable statistical guarantees on resource coverage and deadline satisfaction, without prior knowledge of the pipeline structure or its distribution.

Two complementary strands of related work approach resource provisioning under uncertainty, downstream of service discovery and selection~\cite{farhoudi2025service}.
The first strand, stochastic workflow scheduling for cloud systems, addresses \ac{dag}-structured workloads.
Ye \textit{et al.}~\cite{ye2023stochastic} formulate stochastic multi-workflow scheduling in clouds and derive cost-minimizing decisions under a normal approximation ($\rho = \mathbb{E}[t] + \mathrm{std}(t)$) for task execution times.
EPOSS~\cite{eposs2024} fits per-task Gamma distributions via \ac{mle} and enforces probabilistic deadline constraints, scaling to large \ac{iaas} cloud workloads.
DEFT~\cite{deft2026} trains a \ac{moe} \ac{drl} policy that encodes workflow \acp{dag} and \ac{vm} states for deadline-aware scheduling.

The second strand, resource management for \ac{mec} under uncertainty, addresses joint offloading and bandwidth allocation in hierarchical edge networks via Lyapunov optimization and stochastic programming~\cite{xia_distributed_2024, du_online_2025}, and game-theoretic multi-agent learning for \ac{aigc} services~\cite{liu_joint_2026}.
At the intersection, E$^3$-Agent~\cite{e3agent2026} uses an \ac{llm} meta-controller to adapt routing for edge generative inference under distribution shift, \ac{llm}-based orchestrators combine hierarchical planning with continual reinforcement learning or conversational multi-domain agents~\cite{shokrnezhad2025autonomous, shokrnezhad2026conversational}, and Amayuelas \textit{et al.}~\cite{amayuelas_self-resource_2025} study self-allocation of computational tasks in multi-agent \ac{llm} systems.

The shared limitation of existing work is a structural mismatch with the agentic setting: classical schedulers assume a known pipeline \ac{dag}~\cite{ye2023stochastic, eposs2024, deft2026}; \ac{mec} methods rely on parametric or stationary demand models that do not capture multi-modal, heavy-tailed agentic resource profiles~\cite{xia_distributed_2024, du_online_2025}; \ac{llm}-based orchestrators adapt policies online without dispatch-time coverage guarantees~\cite{e3agent2026, shokrnezhad2025autonomous, shokrnezhad2026conversational}; and the only distribution-free bounds available, namely the raw empirical quantile~\cite{angelopoulos2023gentle} and Chebyshev's inequality, trade one deficiency for another: the former lacks finite-sample coverage guarantees, while the latter over-covers by a large margin.

This paper fills the gap with three contributions.
\textbf{(i)}~We formalize the \emph{uncertainty-aware agentic orchestration problem~(\textbf{P})}, casting compute, bandwidth, and deadline requirements as probabilistic chance constraints over stochastic pipeline totals.
\textbf{(ii)}~We propose \ac{cro}, a conformal prediction-based online reservation algorithm that translates a sliding window of past execution traces into dispatch-time allocations with finite-sample, distribution-free coverage guarantees, requiring no parametric assumption and incurring only the cost of sorting the calibration window at each dispatch.
\textbf{(iii)}~Through simulation over heterogeneous task mixes with multi-modal and heavy-tailed resource profiles, we demonstrate that \ac{cro} is the only method that simultaneously satisfies probabilistic coverage at all target risk levels and, among coverage-feasible methods, attains the lowest over-reservation, roughly halving the resource waste of the Chebyshev bound.

The remainder of this paper is organized as follows.
Section~\ref{sec:problem} formulates the system model and problem.
Section~\ref{sec:method} presents \ac{cro} and its coverage guarantee.
Section~\ref{sec:simulations} provides simulation results, and Section~\ref{sec:conclusion} concludes.

\section{System Model and Problem Formulation}\label{sec:problem}

\subsection{Agentic AI System Model}

\begin{figure}[t]
    \centering
    \includegraphics[width=\columnwidth]{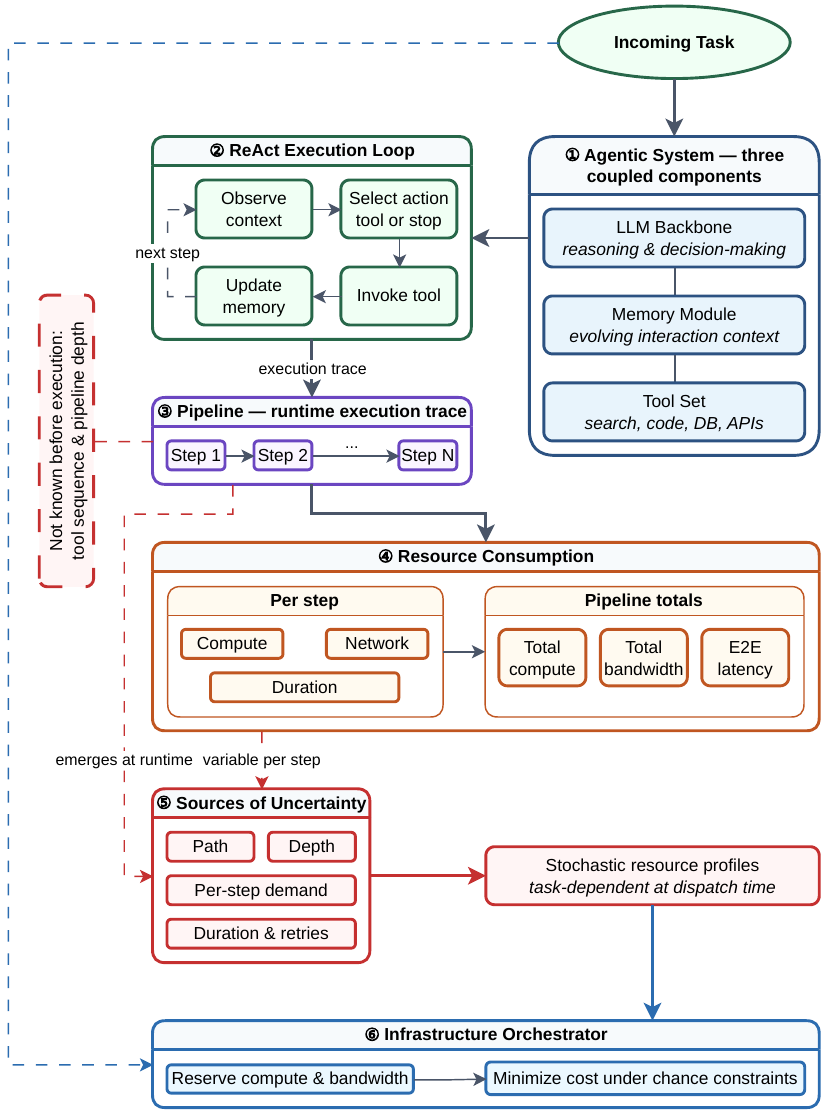}
    \vspace{-10pt}
    \caption{System model overview. The agentic system $\mathcal{A}$
        (LLM backbone, memory, tool set) executes a ReAct loop that
        produces a runtime pipeline $\mathcal{G}=(v_1,\ldots,v_N)$
        whose depth $N$ and tool sequence are unknown at dispatch.
        Each step incurs compute and network demands that aggregate
        into total consumption $C$, $B$, and latency $L$.
        Four structural sources (path, depth, per-step demand, duration)
        make these quantities stochastic, motivating the
        uncertainty-aware orchestration problem~\eqref{eq:P}.}
    \label{fig:system_model}
    \vspace{-10pt}
\end{figure}

We consider an agentic system $\mathcal{A}$ composed of three tightly coupled components:
an \ac{llm} backbone $\mathcal{M}$ responsible for reasoning and decision-making,
a memory module $\mathcal{H}$ maintaining the evolving interaction context,
and a finite tool set $\mathcal{T} = \{t_1, \ldots, t_{|\mathcal{T}|}\}$ of executable
capabilities (e.g., web search, code interpreter, database query, sub-agent call)
\cite{schick2023toolformer, plaat2025agentic}.
Upon receiving a task $q$, the agent executes a \emph{\ac{react}-style} loop
\cite{yao2023react}: at each discrete step $k$, the \ac{llm} observes the
current context $s_k = (q, \mathcal{H}_k)$ and selects an action
$a_k \in \mathcal{T} \cup \{\texttt{stop}\}$.
If $a_k = t_j$, tool $t_j$ is invoked with \ac{llm}-generated parameters,
produces an observation $o_k$, and $o_k$ is appended to $\mathcal{H}$
before the next step. The loop terminates when the agent selects \texttt{stop}
or an execution budget is exhausted \cite{yu2025workflow}.
The resulting execution trace defines a \emph{pipeline}
$\mathcal{G} = (v_1, v_2, \ldots, v_N)$: a sequential chain of $N$ tool
invocations where each step $v_k$ consumes the full accumulated
context $\mathcal{H}_k$ produced by all preceding steps.
Critically, this chain is \emph{not known before execution begins}:
the agent's action at each step is conditioned on intermediate
observations, so both the sequence of invoked tools and the total
length $N$ are determined at runtime.
Figure~\ref{fig:system_model} illustrates the full system model,
from agent components and the ReAct loop through to resource consumption,
uncertainty sources, and the orchestration objective.

\subsection{Resource Consumption Model}

Each tool invocation at step $k$ consumes two classes of infrastructure resources.

\textbf{Compute.}
Every step requires \ac{llm} inference over the accumulated context
$\mathcal{H}_k$ and, if $a_k = t_j$, execution of tool $t_j$
(e.g., running a code interpreter, querying a vector index, or calling a
remote \ac{llm}). Let $c_k \geq 0$ denote the compute demand at step $k$,
measured in normalized units (e.g., \ac{gpu}-seconds).
The value of $c_k$ depends on both the length $|\mathcal{H}_k|$
(which governs \ac{llm} inference cost) and the content of $\mathcal{H}_k$
(which determines the invoked tool and its input parameters,
governing tool-side execution cost) \cite{e3agent2026}.

\textbf{Network.}
Tools that access remote services (\ac{rest}~\acp{api}, cloud \ac{llm} endpoints,
retrieval databases) require network bandwidth.
Let $b_k \geq 0$ denote the average bandwidth demand at step $k$ (Mbps).
Steps invoking local tools have $b_k \approx 0$;
remote tools incur variable $b_k$ depending on payload and response size
\cite{plaat2025agentic}.

The total resource consumption over a pipeline execution is
\begin{equation}
    C = \sum_{k=1}^{N} c_k,
    \qquad
    B = \sum_{k=1}^{N} b_k,
    \label{eq:total_resources}
\end{equation}
and the \ac{e2e} pipeline latency is
\begin{equation}
    L = \sum_{k=1}^{N} \tau_k,
    \label{eq:latency}
\end{equation}
where $\tau_k \geq 0$ is the wall-clock duration of step $k$.

\subsection{Sources of Uncertainty}\label{subsec:uncertainty}

Because $\mathcal{G}$ emerges at runtime, the quantities $C$, $B$, and $L$
in \eqref{eq:total_resources}--\eqref{eq:latency} are \emph{random variables}
from the perspective of the infrastructure orchestrator.
Four structural sources drive this uncertainty:

\begin{enumerate}

    \item \textbf{Path uncertainty.}
          The action $a_k$ selected by the \ac{llm} at each step depends on $s_k$
          and on stochastic token sampling, making $\mathcal{G}$
          unpredictable a priori.
          A step may branch to any $t_j \in \mathcal{T}$, or terminate,
          based on the content of intermediate observations.

    \item \textbf{Depth uncertainty.}
          The total number of steps $N$ is unknown at dispatch time;
          it depends on task complexity, the quality of intermediate results,
          and whether failure-recovery loops (reflection, retry) are triggered
          \cite{plaat2025agentic}.
          Unlike classical workflow scheduling \cite{deft2026, ye2023stochastic},
          the chain length is not known at submission time.

    \item \textbf{Per-step demand uncertainty.}
          Even when tool $t_j$ is invoked, its compute demand $c_k$ and
          bandwidth demand $b_k$ vary with the \ac{llm}-generated input parameters,
          output length, and the current context size $|\mathcal{H}_k|$
          \cite{e3agent2026}.
          Output length is unpredictable, and its effect cascades:
          a longer observation at step $k$ expands $|\mathcal{H}_{k+1}|$,
          affecting $c_{k+1}$.

    \item \textbf{Duration uncertainty.}
          The execution time $\tau_k$ varies due to remote \ac{api} round-trip latency,
          code execution complexity, and \ac{llm} inference time that scales with
          context length \cite{yu2025workflow}.
          Failures and retries convert reliability events into
          additional demand spikes.

\end{enumerate}

Together, these sources yield stochastic resource profiles
$C \sim p_C(\cdot \mid q)$, $B \sim p_B(\cdot \mid q)$, and
$L \sim p_L(\cdot \mid q)$,
where the distributions are task-dependent and partially known
at task-dispatch time.

\subsection{Problem Formulation}

At the moment a task $q$ arrives, the orchestrator must reserve
compute capacity $\rho_c$ (\ac{gpu}-seconds) and network bandwidth $\rho_b$
(Mbps) for the forthcoming pipeline execution.
Under-provisioning risks deadline violations and mid-execution throttling;
over-provisioning wastes scarce edge or cloud resources.
Let $\alpha_c > 0$ and $\alpha_b > 0$ be the unit costs of compute and
bandwidth, $D$ the task deadline, and $C_{\max}$, $B_{\max}$
the available capacity at the serving node.
The \emph{uncertainty-aware orchestration problem} is:
\begin{subequations}
    \label{eq:P}
    \begin{align}
         & \mathbf{(P):}\;\underset{\rho_c,\,\rho_b}{\min}\quad
        \alpha_c \rho_c + \alpha_b \rho_b
        \quad\textit{s.t.}
        \label{eq:obj}                                                    \\
         & \Pr\!\left(L \leq D\right) \geq 1 - \epsilon,
        \tag{C1}\label{C1}                                                \\[2pt]
         & \Pr\!\left(C \leq \rho_c\right) \geq 1 - \delta,
        \tag{C2}\label{C2}                                                \\[2pt]
         & \Pr\!\left(B \leq \rho_b\right) \geq 1 - \delta,
        \tag{C3}\label{C3}                                                \\[2pt]
         & 0 \leq \rho_c \leq C_{\max},\quad 0 \leq \rho_b \leq B_{\max}.
        \tag{C4}\label{C4}
    \end{align}
\end{subequations}
where $\epsilon \in (0,1)$ is the maximum tolerated deadline-violation
probability and $\delta \in (0,1)$ is the maximum probability that
actual resource demand exceeds the reserved allocation.
Constraint~\eqref{C1} requires that the pipeline completes before
deadline $D$ with probability at least $1-\epsilon$,
tolerating a small risk of deadline violation rather than enforcing
an infeasible worst-case guarantee.
Constraints~\eqref{C2} and~\eqref{C3} require that the reserved
compute and bandwidth allocations cover the actual pipeline demand
with probability at least $1-\delta$; any excess demand beyond
$\rho_c$ or $\rho_b$ would cause throttling or queuing at runtime.
Constraint~\eqref{C4} enforces that reservations do not exceed
the available node capacity.

\begin{remark}\label{remark:problem_challenges}
    Problem \eqref{eq:P} is challenging for three
    reasons:
    (i)~the distributions $p_C$, $p_B$, and $p_L$ are not available in closed
    form and must be estimated from execution traces;
    (ii)~$C$, $B$, and $L$ are positively correlated through the shared
    random depth $N$, so the constraints cannot be decoupled;
    and (iii)~the reservation decision must be made online at low
    overhead, ruling out exhaustive Monte Carlo rollouts at runtime.
\end{remark}
\section{Proposed Method}\label{sec:method}

\subsection{Overview}

The three challenges identified in Remark~\ref{remark:problem_challenges} call
for a method that is simultaneously non-parametric, statistically rigorous, and
computationally lightweight.
We propose the \textit{\ac{cro}}, which casts the
probabilistic constraints of problem~\eqref{eq:P} as \emph{conformal prediction
    intervals} computed from a sliding window of past execution traces.
By construction, \ac{cro} inherits the finite-sample, distribution-free coverage
guarantee of conformal prediction~\cite{vovk2005algorithmic,
    angelopoulos2023gentle}: no assumption on the shape of $p_C$, $p_B$, or $p_L$ is needed,
and the guarantee holds for any window size $n \geq 1$, as made precise below.

\subsection{Calibration Window and Nonconformity Scores}

Let $\mathcal{D}_n = \{(C^{(i)},\, B^{(i)},\, L^{(i)})\}_{i=1}^{n}$ be the
\emph{calibration window}: the $n$ most recently completed pipeline traces, each
recording realized total compute, bandwidth, and latency for one task as defined in
\eqref{eq:total_resources}--\eqref{eq:latency}.
The window is capped at a configurable size $W$ (i.e., $n \leq W$),
and traces within the window are treated as exchangeable with the arriving task,
which holds when the task distribution is approximately stationary over the window horizon.
For each trace $i$, the \emph{nonconformity score} for compute is
\begin{equation}
    s_{C}^{(i)} \;=\; C^{(i)} - \hat{C},
    \label{eq:nc_score}
\end{equation}
\noindent where $\hat{C} = \frac{1}{n}\sum_{j=1}^{n} C^{(j)}$ is the rolling sample
mean\footnote{Any predictor~$\hat{C}$ trained on a disjoint fitting set can
    replace the rolling mean without affecting the coverage guarantee; the mean is
    used here for $O(1)$ online updates.}.
Scores $s_{B}^{(i)} = B^{(i)} - \hat{B}$ and
$s_{L}^{(i)} = L^{(i)} - \hat{L}$ are defined analogously.
A large positive score flags a pipeline whose actual demand exceeded the typical
level, capturing the tail behavior driven by depth and per-step demand
uncertainty (sources~2--3 in Section~\ref{subsec:uncertainty}).

\subsection{Conformal Reservations and Feasibility Check}

When a new task $q$ arrives, \ac{cro} uses the scores in $\mathcal{D}_n$
to compute reservations $(\rho_c, \rho_b)$ specifically for~$q$
before dispatching it.

\textbf{Resource reservations.}
For target violation levels $\delta$ (resource overflow) and $\epsilon$
(deadline miss), let $k_{C}=\lceil(1-\delta)(n+1)\rceil$ and
$k_{L}=\lceil(1-\epsilon)(n+1)\rceil$.
If $k_{C}=n+1$ or $k_{L}=n+1$ (i.e., $n<(1-\delta)/\delta$ or
$n<(1-\epsilon)/\epsilon$), that order statistic among $n$ scores does not
exist: we set the corresponding quantile to $+\infty$ and
\textsc{Dispatch} returns \textsc{Infeasible}~\cite{vovk2005algorithmic}.
Otherwise
\begin{equation}
    \hat{q}_{C} = \mathrm{Quantile}\!\left(
    \bigl\{s_{C}^{(i)}\bigr\}_{i=1}^{n};\;
    k_{C}/n
    \right),
    \label{eq:qhat_C}
\end{equation}
with $\hat{q}_{B}$ and $\hat{q}_{L}$ at ranks $k_{C}$ and $k_{L}$.
The unclamped reservations are
\begin{align}
    \rho_c & = \hat{C} + \hat{q}_{C},
    \label{eq:res_c}                  \\
    \rho_b & = \hat{B} + \hat{q}_{B}.
    \label{eq:res_b}
\end{align}

\textbf{Feasibility check.}
\Ac{e2e} latency $L$ cannot be reserved directly, so \ac{cro} checks whether
its $(1-\epsilon)$ conformal bound fits the deadline:
\begin{equation}
    \hat{L} + \hat{q}_{L} \leq D.
    \label{eq:deadline_check}
\end{equation}
\textsc{Dispatch} also returns \textsc{Infeasible} if any quantile is infinite
or if $\rho_c > C_{\max}$ or $\rho_b > B_{\max}$.
Clamping to $C_{\max}$ would replace $\{C \leq \hat{C}+\hat{q}_{C}\}$ by
$\{C \leq C_{\max}\}$, which need not hold with probability $1-\delta$.
An infeasible task may be deferred, re-routed, or rejected by operator policy.

\subsection{Online Window Update and Complexity}

After each pipeline completes, \ac{cro} appends $(C, B, L)$ to $\mathcal{D}$ and
drops the oldest entry once $|\mathcal{D}|$ exceeds $W$ (circular buffer).
Rolling means are updated in $O(1)$; the conformal quantiles require sorting
$n$ scores, costing $O(W \log W)$ per dispatch.
For practical window sizes ($W \leq 10^3$) this is sub-millisecond, making \ac{cro}
suitable for online orchestration without blocking the dispatch path.
Algorithm~\ref{alg:cro} summarizes the full procedure.

\begin{algorithm}[!tb]
    \caption{\acf{cro}}
    \label{alg:cro}
    \begin{algorithmic}[1]
        \Procedure{Dispatch}{$\mathcal{D}_n, D, \epsilon, \delta, C_{\max}, B_{\max}$}
        \State $n \gets |\mathcal{D}_n|$;\;
        $k_C \gets \lceil(1{-}\delta)(n{+}1)\rceil$;\;
        $k_L \gets \lceil(1{-}\epsilon)(n{+}1)\rceil$
        \If{$k_C > n$ or $k_L > n$}
        \State \Return \textsc{Infeasible} \Comment{quantile undefined}
        \EndIf
        \State Compute rolling means $\hat{C}, \hat{B}, \hat{L}$ from $\mathcal{D}_n$
        \State Compute scores $s_C^{(i)}, s_B^{(i)}, s_L^{(i)}$ for all $i$ via \eqref{eq:nc_score}
        \State $\hat{q}_C \gets \mathrm{Quantile}\bigl(\{s_C^{(i)}\};\, k_C/n\bigr)$ \Comment{cf.\ \eqref{eq:qhat_C}}
        \State $\hat{q}_B \gets \mathrm{Quantile}\bigl(\{s_B^{(i)}\};\, k_C/n\bigr)$
        \State $\hat{q}_L \gets \mathrm{Quantile}\bigl(\{s_L^{(i)}\};\, k_L/n\bigr)$
        \State $\rho_c \gets \hat{C} + \hat{q}_C$;\;
        $\rho_b \gets \hat{B} + \hat{q}_B$
        \If{$\hat{L} + \hat{q}_L > D$ or $\rho_c > C_{\max}$ or $\rho_b > B_{\max}$}
        \State \Return \textsc{Infeasible} \Comment{deadline or capacity}
        \EndIf
        \State \Return $(\rho_c,\, \rho_b)$
        \EndProcedure
        \Statex
        \Procedure{OnComplete}{$C, B, L, \mathcal{D}, W$}
        \State Append $(C, B, L)$ to $\mathcal{D}$
        \If{$|\mathcal{D}| > W$} drop oldest entry \EndIf
        \EndProcedure
    \end{algorithmic}
\end{algorithm}

\subsection{Coverage Guarantee}

\begin{theorem}[Marginal coverage]\label{thm:coverage}
    Suppose the traces in $\mathcal{D}_n \cup \{(C, B, L)\}$ are exchangeable,
    where $(C,B,L)$ denotes the random totals of the incoming task~$q$ ($(n+1)$-th task).
    If $k_C \leq n$ and $k_L \leq n$,
    \begin{equation}
        \begin{aligned}
            \Pr\!\bigl(C \leq \hat{C}+\hat{q}_{C}\bigr) & \geq 1 - \delta,   \\
            \Pr\!\bigl(B \leq \hat{B}+\hat{q}_{B}\bigr) & \geq 1 - \delta,   \\
            \Pr\!\bigl(L \leq \hat{L}+\hat{q}_{L}\bigr) & \geq 1 - \epsilon.
        \end{aligned}
        \label{eq:coverage_guarantee}
    \end{equation}
    Whenever \textup{\textsc{Dispatch}} returns a finite reservation
    $(\rho_c, \rho_b)$, these bounds are unclamped
    ($\rho_c = \hat{C}+\hat{q}_{C} \leq C_{\max}$,
    $\rho_b = \hat{B}+\hat{q}_{B} \leq B_{\max}$)
    and $\hat{L}+\hat{q}_{L} \leq D$, so
    $\{L \leq \hat{L}+\hat{q}_{L}\} \subseteq \{L \leq D\}$.
\end{theorem}
\begin{proof}[Proof sketch]
    Let $C_{(k_C)}$ be the $k_C$-th smallest compute value in the
    calibration window. Exchangeability makes the incoming value $C$
    equally likely to have any of the $n+1$ ranks. Thus,
    \[
        \Pr(C \leq C_{(k_C)}) \geq \frac{k_C}{n+1} \geq 1-\delta .
    \]
    Subtracting the common predictor $\hat C$ does not change the order
    of the calibration values. Hence
    $\hat C+\hat q_C=C_{(k_C)}$, yielding the compute bound; the same
    argument gives $\hat B+\hat q_B=B_{(k_C)}$. If $k_C=n+1$, the required
    calibration order statistic is unavailable and \textsc{Dispatch}
    returns \textsc{Infeasible}.
    A finite return uses these unclamped bounds: replacing
    $\hat C+\hat q_C$ by $C_{\max}$ could invalidate the coverage claim.
    Applying the same rank argument to latency gives
    $\Pr(L\leq\hat L+\hat q_L)\geq1-\epsilon$. Finally, a finite return
    satisfies~\eqref{eq:deadline_check}, so
    $\{L\leq\hat L+\hat q_L\}\subseteq\{L\leq D\}$.
\end{proof}

\begin{remark}\label{remark:coverage}
    Theorem~\ref{thm:coverage} provides separate (\emph{marginal})
    guarantees for compute, bandwidth, and latency, as required
    by~\eqref{C2}--\eqref{C3}. For a finite return, the latency bound also
    implies~\eqref{C1}. It does not guarantee that compute and latency are
    simultaneously within their bounds. The union bound gives
    \[
        \Pr(C\leq\rho_c \text{ and } L\leq\hat L+\hat q_L)
        \geq 1-\delta-\epsilon.
    \]
    The tighter product $(1-\delta)(1-\epsilon)$ would require independence,
    which is not expected because both quantities depend on the random
    pipeline depth $N$. Using a rank below $k_C$ loses the finite-sample
    guarantee. We claim no universal nonparametric optimality; among the
    evaluated methods satisfying~\eqref{C2}, \ac{cro} yields the least waste
    in Section~\ref{sec:simulations}.
\end{remark}

\section{Simulation Results}\label{sec:simulations}

\subsection{Setup}

\textbf{Synthetic trace model.}
Because real-world agentic execution logs at scale are not yet publicly available,
we construct a synthetic trace generator that faithfully captures all four
uncertainty sources of Section~\ref{subsec:uncertainty}.
Each pipeline belongs to one of three task types
(\textit{light}, \textit{medium}, \textit{heavy}) sampled with equal probability.
For light and medium tasks the pipeline depth follows
$N \sim 1 + \mathrm{Poisson}(\lambda_j)$, matching the standard stochastic
workflow assumption \cite{ye2023stochastic}.
For heavy tasks we use
$N \sim 1 + \mathrm{NegBin}(r{=}2,\, p{=}1/6)$, which gives the same mean
$\mathbb{E}[N] = 10$ but variance $\mathrm{Var}[N] = 60$ (versus~10 for Poisson),
modeling the high variability in pipeline depth caused by reflection
and retry loops \cite{plaat2025agentic}.
Per-step demands are drawn independently as
$c_k \sim \Gamma(\kappa_c^j, \theta_c^j)$ and
$b_k \sim \Gamma(\kappa_b^j, \theta_b^j)$
(shape and scale), and step latency follows
$\tau_k = s_j c_k + \mathrm{Exp}(e_j)$,
where $s_j$ is a type-specific slope linking inference time to compute load
and $e_j$ is the rate of an independent exponential noise term.
Here $j \in \{\text{light, medium, heavy}\}$ indexes the task type;
all per-type parameters are listed in Table~\ref{tab:params}.
The compound Gamma-NegBin distribution of $C$ for heavy tasks
is positively skewed and heavy-tailed (strictly harder than a single Gamma),
and the equal-type mixture creates a multi-modal marginal
that no single parametric family can approximate well.

\begin{figure*}[t]
    \centering
    \includegraphics[width=\textwidth]{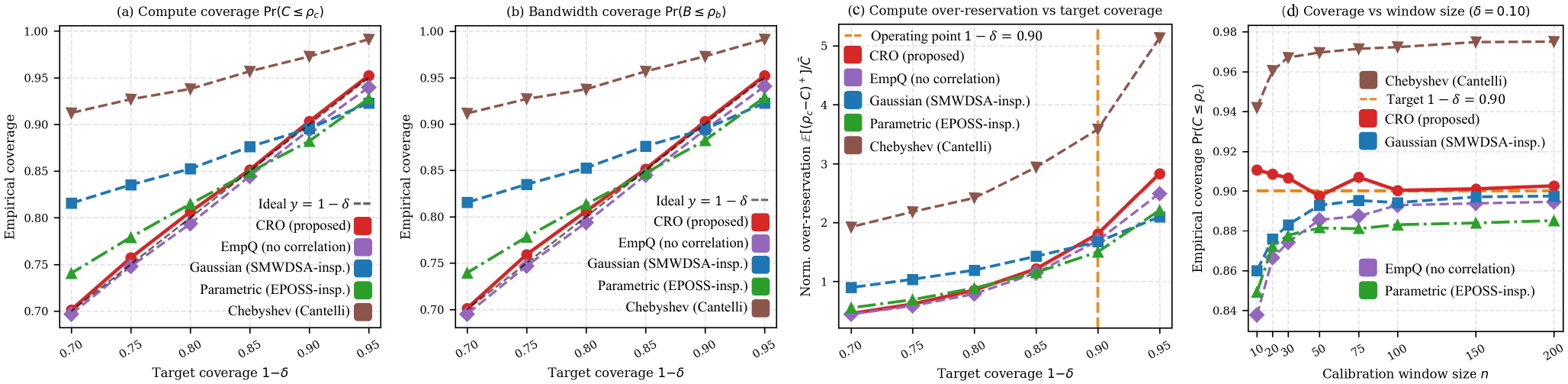}
    \vspace{-10pt}
    \caption{Simulation results ($n{=}100$ unless noted, $\epsilon{=}0.10$, 400 trials).
        (a) Compute coverage $\Pr(C \leq \rho_c)$ vs.\ target $1{-}\delta$;
        (b) bandwidth coverage $\Pr(B \leq \rho_b)$ vs.\ target;
        (c) normalized compute over-reservation vs.\ target coverage;
        (d) compute coverage vs.\ calibration window size at $\delta{=}0.10$.
        \acs{cro} tracks the ideal diagonal in (a)--(b) and meets the $0.90$ target
        in (d) for all $n \geq 10$; EmpQ, Gaussian, and Parametric under-reserve;
        Chebyshev over-covers at nearly $2{\times}$ the waste of \acs{cro} in (c).}
    \label{fig:sim_results}
    \vspace{-10pt}
\end{figure*}

\begin{table}[t]
    \centering
    \caption{Simulation Parameters ($j\!\in\!\{\text{lt, md, hv}\}$)}
    \label{tab:params}
    \scriptsize
    \setlength{\tabcolsep}{3pt}
    \begin{tabular}{lccc}
        \hline
        \textbf{Parameter (symbol)}     & \textbf{Light}                           & \textbf{Medium}       & \textbf{Heavy}                   \\
        \hline
        \multicolumn{4}{l}{\textit{System constants}}                                                                                         \\
        Compute capacity ($C_{\max}$)   & \multicolumn{3}{c}{200 \ac{gpu}-s}                                                                  \\
        Bandwidth capacity ($B_{\max}$) & \multicolumn{3}{c}{100 Mbps}                                                                        \\
        Task deadline ($D$)             & \multicolumn{3}{c}{30 s}                                                                            \\
        \hline
        \multicolumn{4}{l}{\textit{Per-type trace model}}                                                                                     \\
        Depth ($N_j$)                   & $1{+}\mathrm{Poi}(3)$                    & $1{+}\mathrm{Poi}(6)$ & $1{+}\mathrm{NB}(2,\frac{1}{6})$ \\
        Compute shape ($\kappa_c^j$)    & 2.0                                      & 2.5                   & 3.0                              \\
        Compute scale ($\theta_c^j$)    & 0.5                                      & 1.0                   & 1.5                              \\
        Bandwidth shape ($\kappa_b^j$)  & 1.5                                      & 2.0                   & 2.5                              \\
        Bandwidth scale ($\theta_b^j$)  & 0.3                                      & 0.6                   & 0.9                              \\
        Latency slope ($s_j$)           & 0.30                                     & 0.50                  & 0.80                             \\
        Latency noise rate ($e_j$)      & 0.20                                     & 0.40                  & 0.70                             \\
        \hline
        \multicolumn{4}{l}{\textit{Experiment settings}}                                                                                      \\
        Calibration window ($n$)        & \multicolumn{3}{c}{10–200 (default 100)}                                                            \\
        Risk level ($\delta$)           & \multicolumn{3}{c}{0.05–0.30}                                                                       \\
        Deadline tolerance ($\epsilon$) & \multicolumn{3}{c}{0.10}                                                                            \\
        Test tasks / trial              & \multicolumn{3}{c}{1\,000}                                                                          \\
        Monte Carlo trials              & \multicolumn{3}{c}{400}                                                                             \\
        \hline
    \end{tabular}
\end{table}

\textbf{Baselines.}
We compare \ac{cro} against four complementary baselines:
\begin{enumerate}[label=(\roman*)]
    \item \textbf{EmpQ}~\cite{angelopoulos2023gentle}: uses the raw empirical
          $(1{-}\delta)$-quantile with no conformal correction, and the
          standard non-conformity quantile before the finite-sample $+1$
          adjustment of \cite{vovk2005algorithmic} is applied.
          It isolates the exact contribution of the conformalization step in \ac{cro}.
    \item \textbf{Gaussian}~\cite{ye2023stochastic}: sets
          $\rho_c = \hat{C} + \Phi^{-1}(1{-}\delta)\,\hat{\sigma}_C$,
          the Normal-approximation quantile,
          corresponding to $\rho = \mathbb{E}[t] + \mathrm{std}(t)$
          in Ye et al.~\cite{ye2023stochastic}.
          It fails when $C$ is non-Gaussian.
    \item \textbf{Parametric}~\cite{eposs2024}: fits a Gamma distribution via
          \ac{mle} on the calibration window and uses the theoretical
          $(1{-}\delta)$-quantile, following the approach of EPOSS~\cite{eposs2024} adapted online.
          It fails when the distribution is heavier-tailed than Gamma.
    \item \textbf{Chebyshev}: uses the Cantelli one-sided bound
          $\rho_c = \hat{C} + \sqrt{(1{-}\delta)/\delta}\,\hat{\sigma}_C$,
          which is distribution-free and always guarantees coverage,
          but is far more conservative than \ac{cro}.
\end{enumerate}

\textbf{Metrics.}
(1)~\emph{Empirical coverage} $\hat{p}_C = \Pr(C \leq \rho_c)$
and $\hat{p}_B = \Pr(B \leq \rho_b)$ on held-out test traces;
(2)~\emph{normalized over-reservation}
$\mathbb{E}[(\rho_c{-}C)^+] / \bar{C}$;
(3)~\emph{deadline feasibility accuracy}: whether
check~\eqref{eq:deadline_check} correctly identifies $\Pr(L{>}D) \leq \epsilon$.

\subsection{Results}

\textbf{Coverage calibration.}
Fig.~\ref{fig:sim_results}(a)--(b) sweeps $\delta \in \{0.05, \ldots, 0.30\}$
at fixed $n=100$.
\ac{cro} tracks the ideal diagonal ($y = 1{-}\delta$) throughout:
at $\delta = 0.10$ it achieves $\hat{p}_C = 0.903$, confirming
Theorem~\ref{thm:coverage}. EmpQ runs nearly parallel to \ac{cro} but sits consistently
below the diagonal ($\hat{p}_C = 0.894$ at $\delta = 0.10$).
The gap is small at $n=100$ because the conformal correction shifts only one
quantile rank; yet it is systematic and grows at smaller $n$ (see below),
confirming that the $+1$ adjustment is non-trivial for finite samples.
Gaussian exhibits a crossing pattern: it over-covers at
$1{-}\delta \leq 0.80$ and under-covers at $1{-}\delta \geq 0.85$,
reaching $\hat{p}_C = 0.919$ at target $0.95$, a violation
of constraint~\eqref{C2} exactly when \acp{sla} are most stringent.
Parametric under-covers monotonically ($1.8\% - 4.3\%$ deficit),
with the Gamma \acs{mle} fit consistently missing the compound Gamma-NegBin upper tail.
Chebyshev always exceeds the diagonal
($\hat{p}_C = 0.973$ at $\delta = 0.10$), proving coverage but at
the cost of a normalized over-reservation of $3.58$, nearly $2{\times}$
that of \ac{cro} (Fig.~\ref{fig:sim_results}(c)).

\textbf{Resource efficiency and window size.}
Fig.~\ref{fig:sim_results}(c) shows that among the five compared methods only
\ac{cro} simultaneously meets the coverage constraint and minimizes waste.
At $\delta = 0.10$: EmpQ ($1.69$), Gaussian ($1.67$),
and Parametric ($1.50$) all appear cheaper than \ac{cro} ($1.81$),
but they achieve this by under-reserving, violating constraint~\eqref{C2}.
Chebyshev (waste $= 3.58$) is the only other method that provably
meets the constraint, but at $2{\times}$ the over-reservation of \ac{cro}.
The $0.14$ additional normalized waste of \ac{cro} over EmpQ is the finite-sample
correction cost.
Among methods that meet~\eqref{C2}, \ac{cro} has the least waste here; we do not claim optimality over all nonparametric procedures.
Fig.~\ref{fig:sim_results}(d) shows the finite-sample advantage most
dramatically.
At $n = 10$, \ac{cro} already achieves $\hat{p}_C = 0.911$;
EmpQ drops to $0.838$, a $7.3$ percentage-point deficit, because without the
conformal correction there is no finite-sample guarantee on the empirical
quantile.
Gaussian ($0.860$) and Parametric ($0.849$) also start
well below target and never recover: both plateau permanently below $0.90$.
Chebyshev, conversely, over-covers at all $n$ ($\hat{p}_C > 0.94$),
confirming that its conservatism is not a data-scarcity artifact.

\textbf{Deadline feasibility check.}
At $\epsilon = 0.10$, \ac{cro}'s feasibility test~\eqref{eq:deadline_check}
achieves perfect accuracy across all 400 trials: it correctly flags
heavy-type pipelines as infeasible despite $\bar{L} < D$, because
NegBin-depth tail latency drives $\Pr(L{>}D)$ above $10\%$.
All quantile-based baselines (EmpQ, Gaussian, Parametric, Chebyshev)
also achieve perfect feasibility accuracy here, confirming that
tail-aware latency reservation is necessary and sufficient for
the deadline check, but only \ac{cro} simultaneously provides the
finite-sample guarantee on $C$ and $B$ with minimal waste.
The plotted grid has $k_C,k_L \leq n$; the capacity test is inactive except in
fewer than $2\%$ of $n{=}10$ trials.

\section{Conclusion}\label{sec:conclusion}

This paper addressed the problem of dispatch-time resource reservation for agentic \ac{ai} pipelines, whose aggregate compute, bandwidth, and latency demands are unknown at dispatch time and emerge at runtime as heavy-tailed, multi-modal random variables.
We formalized this as the uncertainty-aware agentic orchestration problem and proposed \ac{cro}, an online algorithm grounded in conformal prediction.
CRO maintained a sliding window of past execution traces and derived per-task resource reservations with finite-sample, distribution-free coverage guarantees, requiring no parametric assumption on the pipeline distribution and incurring only $O(W \log W)$ overhead per dispatch.
Simulations over heterogeneous task mixes confirmed that CRO was the only evaluated method that simultaneously satisfied target probabilistic coverage across all risk levels while, among coverage-feasible methods, attaining the lowest over-reservation, roughly halving the resource waste of the Chebyshev-bound baseline and remaining reliable from as few as ten calibration traces.

Several directions remain open for future work.
First, extending CRO to jointly reserve resources for batches of concurrent tasks would reduce over-reservation through cross-task diversification.
Second, incorporating distribution-shift detection into the window update would strengthen the exchangeability assumption under non-stationary workloads.
Third, validating the approach on real-world agentic logs and integrating it with production orchestration platforms such as Kubernetes, as well as with \ac{llm}-based network orchestrators~\cite{shokrnezhad2025autonomous, shokrnezhad2026conversational}, represent important steps toward practical deployment.
Fourth, coupling CRO with GenAI-driven service discovery~\cite{farhoudi2025service} and with key-value objectives extracted from service descriptions~\cite{shokrnezhad2026kpi2kvi} would close the loop from service intent to coverage-guaranteed allocation.

\section*{Acknowledgment}
This work was, in part, supported by BMFTR, Germany (6GEM+, Grant 16KIS2411), and the EU Horizon Europe programme (6G-Path, Grant 101139172).

\bibliographystyle{IEEEtran}
\bibliography{main}

\end{document}